\documentclass{ecai} 

\usepackage{latexsym}
\usepackage{amssymb}
\usepackage{amsmath}
\usepackage{amsthm}
\usepackage{booktabs}
\usepackage{enumitem}
\usepackage{graphicx}
\usepackage{color}

\newcommand{\BibTeX}{B\kern-.05em{\sc i\kern-.025em b}\kern-.08em\TeX}

\begin{document}


\begin{frontmatter}




\title{NeuralFlix: Reconstructing Vivid Videos from Human Brain Activity}


\author[A]{Jingyuan Sun \footnote{Equal contribution.}}
\author[A]{Mingxiao Li \footnote{Equal contribution.}}
\author[B]{Zijiao Chen} 
\author[A]{Marie-Francine Moens}
\address[A]{Department of Computer Science, KU Leuven, Leuven, Belgium}
\address[B]{School of Medicine, National University of Singapore, Singapore}


\begin{abstract}
In our quest to decode the visual processing of the human brain, we aim to reconstruct dynamic visual experiences from brain activities, a task both challenging and intriguing. Although recent advances have made significant strides in reconstructing static images from non-invasive brain recordings, the translation of continuous brain activities into video formats has not been extensively explored. Our study introduces NeuralFlix, a novel dual-phase framework designed to address the inherent challenges in decoding fMRI data, such as noise, spatial redundancy, and temporal lags. The framework employs spatial masking and temporal interpolation for contrastive learning of fMRI representations, and a diffusion model enhanced with dependent prior noise for generating videos. Tested on a publicly available fMRI dataset, NeuralFlix demonstrates promising results, significantly outperforming previous state-of-the-art models by margins of \(20.97\%\), \(31.00\%\), and \(12.30\%\), respectively, in decoding the brain activities of three subjects, as measured by SSIM. Furthermore, our attention analysis indicates that the model's outputs align with known brain structures and functions, underscoring its biological plausibility and interpretability.
\end{abstract}

\end{frontmatter}


\section{Introduction}

Human experiences are inherently fluid, evolving continuously like scenes within a film \cite{varela2017embodied,chen2023cinematic}. The human brain, remarkable in its complexity, incessantly processes these visual inputs, creating a complex narrative of perceptions \cite{bartels2008natural,nishimoto2011reconstructing}. Deciphering the layers of this intricate system to comprehend the underlying neural processes presents a formidable task. A significant ambition in this realm is to decode from brain activity using non-invasive techniques to reconstruct  human visual perceptions\cite{kupershmidt2022penny, wang2022reconstructing}. Such advancements could largely enhance accessibility for those with sensory impairments and enriching our grasp of the neural foundations of visual perception.

Functional Magnetic Resonance Imaging (fMRI) a non-invasive neuro-imaging technology well-regarded for its high spatial resolution, which is crucial for recording detailed activations in the visual cortex and other brain regions \cite{henaff2021primary}. This capability is essential for reconstructing visual content based on brain activity. While reconstructing static images has seen notable progress \cite{chenzj,sun2023decoding}, reconstructing videos is still an area of active research. One of the main challenges with fMRI is its reliance on measuring changes in the blood-oxygen-level-dependent (BOLD) signal, which can exhibit spatial redundancy and sometimes lag behind actual neural activity due to the hemodynamic response \cite{uugurbil2013pushing, de2009hemodynamic}. Additionally, the non-invasive nature of fMRI means it can pick up noise from various physiological and scanner-related sources, which complicates the reconstruction of high-quality videos \cite{parrish2000impact}.

To address these challenges, we have developed a two-phase framework called NeuralFlix, which aims to reconstruct high-resolution videos from fMRI data. In the first phase, we use spatial masking and temporal interpolation to enhance fMRI data, while an optimized fMRI encoder is trained to resist disturbances from these augmentations. In the second phase, this trained fMRI encoder guides a video diffusion model in generating videos. We further improve this phase by introducing noise models that compensate for the low signal-to-noise ratio typical of fMRI data. Together, these innovations help our framework to convert complex and noisy fMRI data into precise and meaningful visual reconstructions, showcasing the potential of combining advanced neural imaging with machine learning to decode brain activity.

We tested our method on a publicly available dataset of fMRI-video pairs, involving three individuals watching videos. Our results show significant improvements over previous models in both detailed pixel accuracy and broader semantic understanding. Specifically, our approach improves upon the latest state-of-the-art model \cite{chen2023cinematic} by significant margins: 20.97\% in decoding brain activities of Subject 1, 31.00\% in Subject 2, and 12.30\% in Subject 3. Additionally, an analysis of how our model focuses attention shows alignments with both the visual cortex and higher cognitive networks, suggesting our method is not only effective but also biologically plausible. These findings highlight the potential of our approach to advance the fields of neural decoding and visual reconstruction.

\section{Related Works}

\subsection{Decoding Visual Contents from Brain Activities}

\subsubsection{Reconstructing Images from Brain Activities}

Recent advancements in deep generative models have spurred significant interest in the field of reconstructing visual content from brain activities, focusing on both viewed and imagined images \cite{Fang2021ReconstructingPI}. Early research predominantly transformed fMRI signals into image features, which were then processed by fine-tuned Generative Adversarial Networks (GANs) to create images \cite{Mozafari2020ReconstructingNS}. A notable example is the use of a pre-trained VGG network to extract hierarchical image features from fMRI data, which were then used to synthesize images through a GAN \cite{shen2019end}. More recent efforts in the last couple of years have shifted towards employing Diffusion Models. These models have successfully produced images that are both semantically coherent and visually more accurate \cite{qian2023semantic,lin2022mind, chenzj, sun2023decoding}. For instance, \cite{sun2023contrat} significantly improved fMRI representation learning through denoising techniques and leveraged pixel-level guidance from image auto-encoders to effectively isolate vision-related neural activities from non-relevant noise.

\subsubsection{Reconstructing Videos from Brain Activities}

While advancements in reconstructing static images from brain activities are ongoing, decoding videos from fMRI data remains a significant challenge. Traditional video reconstruction methods from fMRI, which treat the process as a sequence of individual image reconstructions, often result in lower frame rates and inconsistent quality \cite{wen2018neural}. These methods have successfully decoded basic image features and categorical data from fMRI responses to video stimuli. An advancement by \cite{wang2022reconstructing} involved using a linear layer to encode fMRI data and a conditional video GAN to enhance video frame quality and consistency. However, the effectiveness of this technique is constrained by the limited size of available datasets, which is a major limitation given the high data demands of training GANs.

Building on this, \cite{kupershmidt2022penny} developed a separable autoencoder that supported self-supervised learning, yielding better results than \cite{wang2022reconstructing} but still falling short in visual quality and semantic accuracy. Further advancing the field, \cite{chen2023cinematic} introduced contrastive learning and spatial-temporal attention mechanisms to improve fMRI representation accuracy. While their approach also emphasized the spatial-temporal characteristics of fMRI, it did not train the model to handle disruptions in space and time and relied on a basic diffusion model setup without addressing the noisy nature of fMRI data. In this paper, the methods from \cite{kupershmidt2022penny, chen2023cinematic, wang2022reconstructing} serve as baselines for comparison with our enhanced approach.

\subsection{Image and Video Generation with Diffusion Models}

Diffusion Models, inspired by nonequilibrium thermodynamics, are probabilistic models that convert data into Gaussian noise and then reconstruct the original data, demonstrating excellent performance in content generation tasks such as text-to-image \cite{stable_diffusion} and 3D object creation \cite{poole2022dreamfusion}. The typically iterative process, requiring hundreds of steps, has been streamlined by the Denoising Diffusion Implicit Model (DDIM) \cite{ddim}, which decreases the steps needed for high-quality output. Enhancements such as the integration of ordinary differential equation solvers \cite{dpmsolver,dpmsolver2, pseudo}, variance optimization \cite{bao2022analytic}, reduction of exposure bias \cite{li2023alleviating,ning2023elucidating,ning2023input}, and improved noise schedulers \cite{improved_ddpm} have further accelerated inference and enhanced generative quality. Early uses of Diffusion Models for video generation were led by the introduction of the 3D diffusion UNet by VDM \cite{vdm}. Later, ImageN \cite{imagen} developed a cascaded sampling framework with super-resolution techniques for producing high-resolution videos. Additional advances include temporal attention mechanisms by Make-A-Video \cite{makevideo} and integration within latent diffusion models by MagicVideo \cite{zhou2022magicvideo} and LVDM \cite{lvdm} to enhance video generation. In this work, we modify the image diffusion model to support video by adding a temporal layer after each spatial layer and incorporating dependent noises.

\begin{figure*}[ht]
\centering
\small
\includegraphics[width=6.8in]{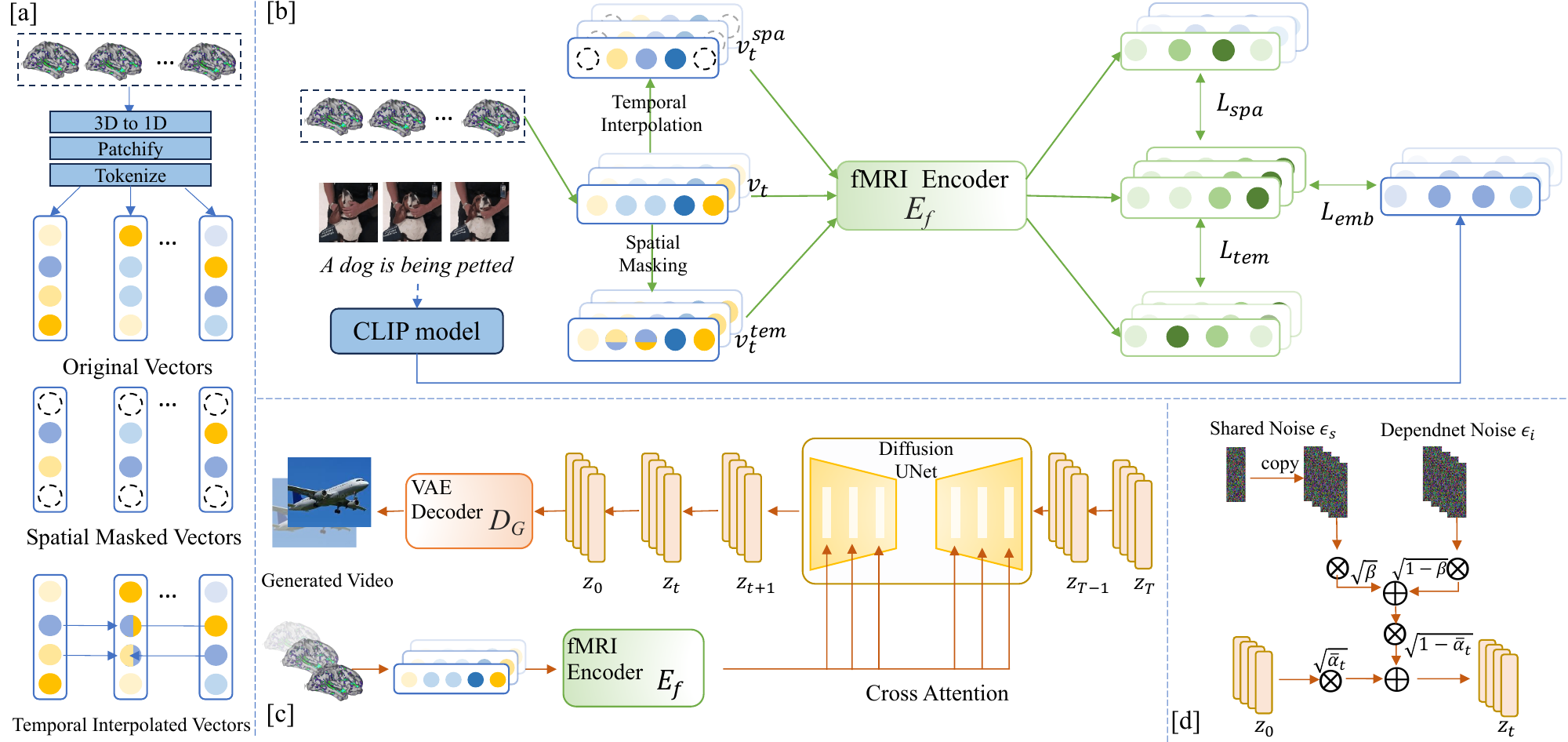}
\caption{The proposed double-phase framework to reconstruct seen videos from fMRI. [a]: Using spatial masking and temporal interpolation to produce augmented samples [b]: Phase 1 trains an fMRI encoder to map from fMRI to CLIP text and image embeddings with contrastive learning. [c]: Phase 2 conditions generation of a diffusion model with Phase 1's fMRI encoder incorporating dependent noises. [d]: Dependent noise generation.} 
\label{main}
\end{figure*}

\section{Method}

Our method consists of fMRI feature leaning and video decoding two-phase framework for reconstructing videos from fMRI-recorded brain activities. Phase 1 involves tuning a pre-trained fMRI encoder with spatial and temporal augmented contrastive learning to align fMRI data with CLIP's text and image features, enhancing the extraction of semantic information from fMRI signals. Phase 2 uses the trained fMRI encoder to guide a video diffusion model, incorporating dependent prior noise to compensate for fMRI's low signal-to-noise ratio.

In Section 3.1, we delve into learning fMRI representation with spatial and temporal augmented contrastive learning. In Section 3.2, we elaborate on the design of prior noise for the video diffusion model to decode more coherent videos from brain activity. In Section 3.3, we describe our experimental approach for analyzing our results, aiming to clarify the contribution of each brain region throughout different stages of learning.

\subsection{FMRI Feature Learning}

\subsubsection{Pre-training and FMRI Input Format}

Given the limited availability of fMRI-video pair data, we leverage a pre-trained fMRI representation space using a model proven effective in image reconstruction from fMRI, as introduced by \cite{sun2023contrat}. This model uses a Vision Transformer-based encoder to process masked fMRI signals and a decoder for restoring unmasked signals, utilizing a double-contrastive masked autoencoding (DC-MAE) technique, optimized on the HCP dataset \cite{hcp}. The "Double-Contrastive" model optimizes contrastive losses through two contrasting operations during fMRI representation learning \cite{sun2023contrat,chen2023cinematic,chenzj, sun2023decoding}. We use this fMRI encoder to establish a robust pre-trained representation space.

For video decoding, we process fMRI data as a sequence of 3D tensors but convert this data into 1D to align with the visual processing hierarchy, segment it into uniform patches, and tokenize, simplifying the model structure due to the scarcity of fMRI-video pairs. To handle the inherent temporal delay in fMRI data, we employ a sliding window technique, defined as $\boldsymbol{v}_{\boldsymbol{t}}=\left\{v_t, v_{t+1}, \ldots, v_{t+w-1}\right\}$, where $v_t \in \mathbb{R}^{n \times p \times b}$ and \( n \), \( p \), and \( b \) represent the batch size, patch size, and embedding dimension, respectively, resulting in $\boldsymbol{v}_{\boldsymbol{t}} \in \mathbb{R}^{n \times w \times p \times b}$, with \( w \) as the window size. Our methods take into account the spatiotemporal characteristics of fMRI data, as further explained in subsequent sections.

\subsubsection{Spatial and Temporal Augmentation}

To tackle the dual challenges of limited fMRI-video pair data and the inherently low signal-to-noise ratio in fMRI, we propose a novel method for training a noise-robust fMRI encoder. Central to our approach is the cognitive plausible augmentation of samples for contrastive learning, tailored to the unique spatial and temporal characteristics of fMRI data. Conventional computer vision augmentations such as cropping and rotation are not suitable for fMRI images, which require maintaining the spatial integrity that correlates with neurological function. After extensive review \cite{glover2011overview,buxton2009introduction} and consultations with experts, we employ two primary augmentation techniques: spatial masking and temporal interpolation.

For spatial masking, we randomly select a portion of the tokens in $\boldsymbol{v}_{\boldsymbol{t}} \in \mathbb{R}^{n \times w \times p \times b}$ and set them to zero. Specifically, $\gamma_{spa}b$ values are zeroed out in the fourth dimension, b, with $\gamma_{spa}$ as a tunable hyperparameter. The positions to be masked are consistent within the same window but vary across different batches.

Temporal interpolation involves replacing randomly selected frames within a window with interpolations of other frames based on their temporal proximity—the farther away a frame, the less it contributes. This method uses weighted interpolations for fMRI sequences, unlike static image augmentation methods like CutMix which involve cropping and pasting. Mathematically, for a window of \( w \) fMRI frames $\boldsymbol{v}_{\boldsymbol{t}}$, and a selected \( i^{th}\) frame ${v_{t_i}}$, the interpolated frame \( \hat{v_{t_i}} \) is calculated as:
\begin{equation}
\hat{v_{t_i}} = \sum\limits_{j=1, j \neq i}^{n} \left (1 - \frac{\lvert i-j \rvert }{n}\right )v_{t_{j}}
\end{equation}
The original frame $v_{t_i}$ is then replaced by $\hat{v}_{t_i}$.

The degree of interpolation is governed by the temporal interpolation ratio $\gamma_{tem}$, another adjustable hyperparameter. Details on the settings and impacts of $\gamma_{tem}$ and $\gamma_{spa}$ will be further explored in the Section 5.2 Ablation Study.

\subsubsection{Contrastive Mapping}

In our method, we employ a vision-Transformer-based fMRI encoder to process fMRI token vectors, aligning them with the CLIP model's text and image embeddings. This process is enhanced by incorporating contrastive learning using augmented examples as described in the previous subsection.

Initially,  we utilize the pre-trained CLIP model \cite{radford2021learning} to encode the stimuli used in the collection of fMRI data. For each video, captions are generated using the pre-trained BLIP model \cite{li2022blip} and then encoded with the CLIP model to produce text embeddings. Similarly, we process each video frame to generate corresponding image embeddings. The fMRI encoder is fed with the token vectors and trained to map from fMRI to CLIP embeddings. Additionally, the fMRI encoder is fed with both the spatially and temporally augmented examples and is optimized with contrastive losses to learn fMRI features robust to the spatial and temporal disturbances.

The formal representation of our loss functions, considering the original fMRI token vectors \( \boldsymbol{v}_{\boldsymbol{t}}\), their spatially augmented version \( \boldsymbol{v}_{\boldsymbol{t}}^{spa} \), and temporally augmented version \( \boldsymbol{v}_{\boldsymbol{t}}^{tem} \), is as follows:
\begin{equation}
\begin{aligned}
    L_{spa} =  &L_{CE}[E_f(\boldsymbol{v}_{\boldsymbol{t}}), E_f(\boldsymbol{v}_{\boldsymbol{t}}^{tem})] \\
    L_{tem} =  &L_{CE}[E_f(\boldsymbol{v}_{\boldsymbol{t}}), \:E_f(\boldsymbol{v}_{\boldsymbol{t}}^{spa})] \\
    L_{emb} =  &L_{CE}[E_f(\boldsymbol{v}_{\boldsymbol{t}}),\: e^{txt}_{t}] +  L_{CE}[E_f(\boldsymbol{v}_{\boldsymbol{t}}),\: e^{img}_{t}],\\
\end{aligned}
\end{equation}

where $L_{CE}$ is the cross-entropy loss and $E_f$ denotes the fMRI encoder. $e^{txt}_{t}$ and $e^{img}_{t}$ mean the CLIP text and image embeddings.  We aim to optimize these losses jointly, with the overall loss function being defined as:
\begin{equation}
L_{E_{f}} = \mu_{spa} L_{spa} + \mu_{tem} L_{tem} + L_{emb}
\end{equation}
In this equation, \( \mu_{spa} \) and \( \mu_{tem} \) are hyperparameters that adjust the weight of the corresponding losses. The setting and effects of
\( \mu_{spa} \) and \( \mu_{tem} \) will be detailed in Section 5.2.


\subsection{Generation with Diffusion Model}
\subsubsection{Prelimiaries} Diffusion Models~\cite{ddpm_ori} show significant potential in generating both images and videos. In this work, we adopt the widely used Stable Diffusion (SD)~\cite{stable_diffusion} as the baseline model, known for its efficient denoising capabilities in the image's latent space, which requires considerably fewer computational resources. During training, the SD begins by using a KL-VAE~\cite{vqvae} encoder to convert image $x_0$ to latent space: $z_0 = \mathcal{E}(x_0)$. It then progressively transforms this latent representation into a Gaussian noise, following the equation:

\begin{equation}
    z_t = \sqrt{\Bar{\alpha}_t}z_0+\sqrt{1-\Bar{\alpha}_t}\epsilon
\end{equation}

Here, the $\epsilon$ represents a noise sampled from a normal distribution: $\epsilon\sim\mathcal{N}(0,1)$. $\Bar{\alpha}_t$ is the predefined noise schedule. The model is trained to predict the added noise at each step, and the loss function could be formulated as :

\begin{equation}
   \mathcal{L}_t^{simple} = E_{t,x_0,\epsilon_t\sim\mathcal{N}(0,1)}[\|\epsilon_t - \epsilon_{\theta}(z_t,t,c)\|_2^2]
\end{equation}

$t$ is the diffusion time step, and $c$ is the text prompt condition. During inference, the SD gradually reconstructs an image aligned with the provided text prompt from Gaussian noise. The denoised results are then processed through the decoder of the KL-VAE to reconstruct the colored images from their latent representation: $x_0 = \mathcal{D}(z_0)$.

\subsubsection{Dependent Prior Video Diffusion.}

Following previous studies \cite{tuneavideo, chen2023cinematic}, we utilize a pre-trained text-to-image Stable Diffusion (SD) model as our foundational video generator. While adept at creating high-quality individual frames, the original SD model lacks temporal coherence for video generation. To address this, we modify it by converting 2D convolutions to pseudo 3D and adding a temporal attention layer after each spatial self-attention layer. This modification introduces temporal awareness, allowing each visual token to attend to tokens from the previous two frames. The temporal attention layer operates as:
\begin{equation}
  Attention(Q, K, V) = Softmax(\frac{QK^T}{\sqrt{d_k}})V  
\end{equation}

with $Q, K, V$ being the query, key, and value matrices, and $W^Q,W^K,W^V$ as learnable parameters.

For decoding brain activity into video, we start by sampling $m$ latent codes from Gaussian noise and progressively refine them using the fMRI representation. Given the low signal-to-noise ratio in fMRI signals, enhancing video quality is challenging. Previous research~\cite{ge2023preserve} demonstrates that employing a deterministic ODE solver in the generative process of the SD model results in a high correlation of initial noise in frames from the same video. Similarly, it has been observed that fMRI signals from similar visual stimuli exhibit a high degree of correlation. 
Based on these observations, we utilize correlated noise as a form of prior knowledge within the generative model and the fMRI decoding process. To create a sequence of dependent noise, where each noise is sampled from Gaussian Distribution with a mean of zero and a variance of one, we divide each noise into two components: $\epsilon_{s}$ and $\epsilon_{i}^j$, and the dependent noise is obtained by following formula: 
\begin{equation}
    \epsilon^j = \sqrt{\beta}\cdot\epsilon_{s} + \sqrt{1-\beta}\cdot\epsilon_{i}^j
\end{equation}
where $\epsilon_s\sim\mathcal{N}(0,1)$ and $\epsilon_i\sim\mathcal{N}(0,1)$. $\sqrt{\beta}$ is the hyperparameter conditioning the noise ratio, whose setting and effects are discussed in Section 5.2's Ablation Study.
A visualization of generating dependent noise is presented in Figure~\ref{all_roi} [d]. During training, we substitute the original noise in SD model with our customized dependent noise to generate noisy latent codes at each time step. Conversely, in the generative phase, the process begins with the introduction of our dependent noise.


\subsection{Interpretation of Brain Activity}


Understanding the activated brain activation patterns is a crucial aspect of our study in brain encoding and decoding. Here, we employ the self-attention mechanism of our fMRI encoder to analyze brain activity in response to various fMRI-video pairs. The networks analyzed are derived from the Yeo17 network~\cite{yeo2011organization}, which includes the visual cortex networks (central and peripheral visual fields, referred to as VisCent and VisPeri), dorsal attention networks (DorsAttnA and DorsAttnB), and the default mode network (DefaultA). The attention maps generated during this process are visualized at different stages of model training, across various layers, and in response to distinct visual stimuli. This approach helps illuminate the contribution of each brain region throughout different learning phases. 

Initially, our analysis begins at the sample level, where self-attention maps are derived from three critical layers within the fMRI encoder, which are the early layer (first layer), the middle layer (12th layer), and the final layer (24th layer). We then average these maps across all video samples to perform a group-level analysis. To correlate the attention values with specific anatomical regions, we reconstruct the fMRI data from its patchified form back into voxel format via the inverse operation of the FC layer. These voxels are then projected onto selected Regions of Interest (ROIs) on the CIFITI den-91k brain surface
~\cite{hcp}. For visualization, we utilize Nilearn~\cite{nilearn} to map these brain voxels back onto the brain surface, enhancing our insights into complex brain activities. Furthermore, we conduct two-sample t-tests comparing the brain activation patterns across different networks and different learning stage in the fMRI encoder. 

\section{Experimental Setup}

\subsection{Evaluation Metrics and Baselines}

Our evaluation metrics are divided into pixel-level and semantics-level assessments. For pixel-level, we use the Structural Similarity Index Measure (SSIM) \cite{wang2004image}. For semantics-level, we employ an N-way top-K accuracy classification test, specifically 50-way-top-1 accuracy. Both SSIM and classification accuracy are computed for each frame against its groundtruth frame. The classification test involves an ImageNet classifier comparing the classifications of the groundtruth and predicted frames. Success is defined when the groundtruth class ranks within the top-K probabilities of the predicted frame's classification from N randomly chosen classes, including the groundtruth itself. This test is repeated 100 times to calculate an average success rate. For video-based metrics, a similar classification approach is used but with a video classifier. This classifier, based on VideoMAE \cite{tong2022videomae} and trained on the Kinetics-400 dataset \cite{kay2017kinetics} for action  classification, assesses video semantics and dynamics across 400 categories, including various motions and human interactions.  We compare our methodology with prior works based on these metrics, including studies from \citet{chen2023cinematic}, \citet{wen2018neural}, \citet{wang2022reconstructing}, and \citet{kupershmidt2022penny}. We cited their results directly from previous SOTA \citet{chen2023cinematic}'s reporting to conduct a fair evaluation.

\subsection{Dataset}

Our study employs a publicly available fMRI-video dataset \cite{wen2018neural} which includes fMRI data from three participants and video clips. The data were collected using a 3T MRI scanner with a 2-second repetition time. The training set consists of 18 video clips, each 8 minutes long, for a total of 2.4 hours, yielding 4,320 paired training examples. The test set comprises five 8-minute video clips totaling 40 minutes and 1,200 test fMRI scans. The videos, displayed at 30 FPS, cover diverse themes like animals, humans, and natural landscapes. There is no complete overlap of semantic categories between the training and testing sets, with a measured overlap of 0.56 based on the intersection\_set/union\_set metric. We reserve 20\% of the training data for validation. For comparability with previous SOTA \cite{chen2023cinematic}, we modify the original videos to a frame rate of 3 FPS and set the window size to 2, translating each fMRI frame to six video frames. This configuration allows for the reconstruction of a 2-second video segment from each fMRI frame, with potential for longer sequences dependent on available GPU memory.

\subsection{Implementation Details}

In the first phase, we use a Vision Transformer (ViT)-based fMRI encoder pre-trained on large-scale fMRI data \cite{sun2023contrat}. The encoder is initially pre-trained with a mask ratio of 0.75 and a patch size of 16, across 24 layers, and embedding dimension of 1024. It also includes a projection head transforming token embeddings into a 77×768 dimensionality. In the second phase, we employ Stable Diffusion V1-5 \cite{stable_diffusion}, fine-tuned to process videos at 256x256 resolution and 3 FPS frame rate. This involves modifying spatial attention, cross-attention, and introducing a temporal layer. Training involves 1000 steps with a learning rate of $2 \times 10^{-5}$ and batch size of 14. Post-training, the text encoder of the video diffusion model is replaced with our fMRI encoder. Further fine-tuning is carried out for spatial self-attention, visual-fMRI cross-attention, and temporal attention, using a learning rate of $2\times 10^{-5}$ and a batch size of 24 on a single A100 GPU.



\begin{figure}[h]
\centering
\small
\includegraphics[width=3.2in]{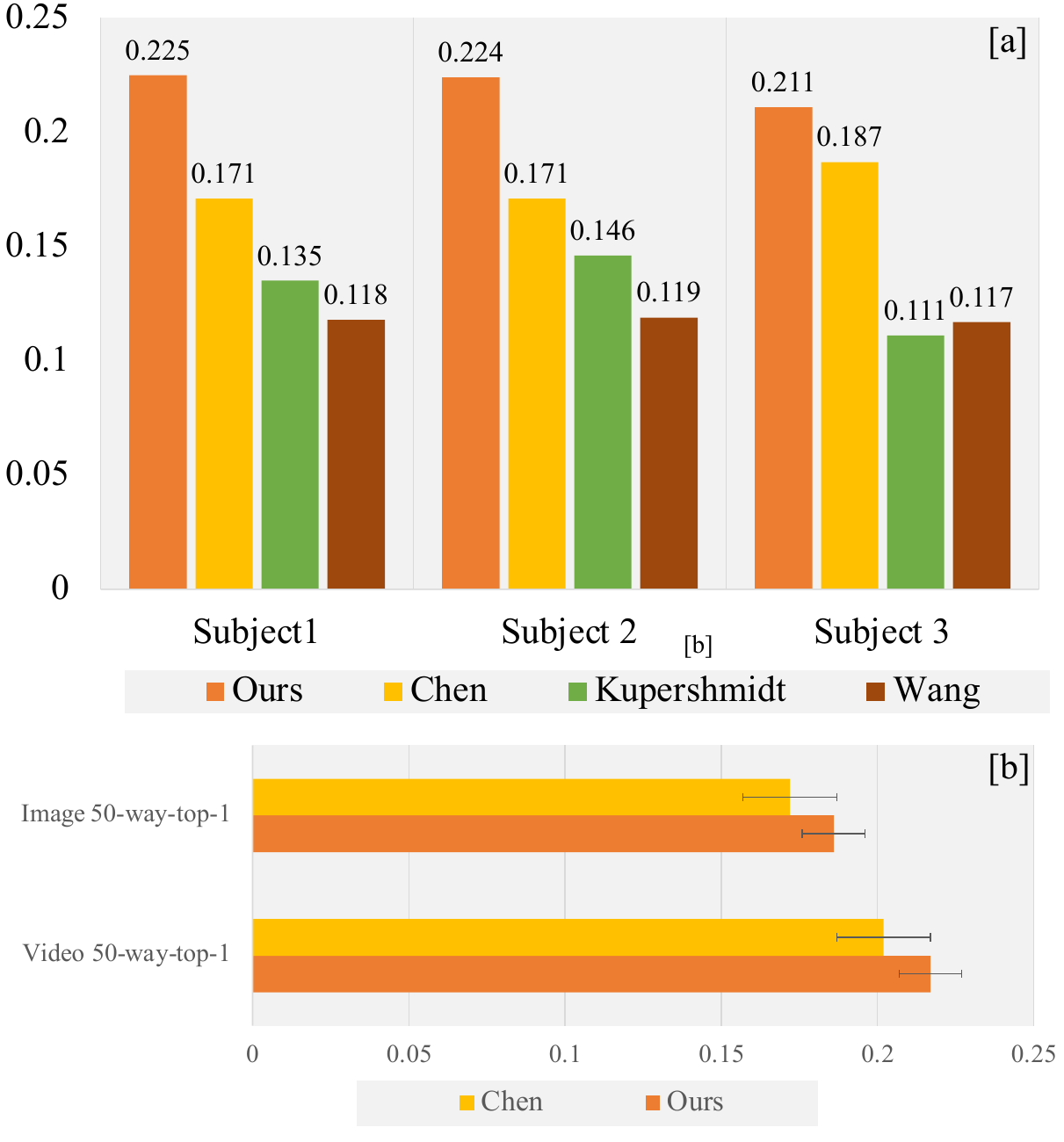}
\caption{Comparisons of Structural Similarity Index Measure (SSIM) Scores and 50-way-top-1 Image/Video Classification
Accuracy.  [a] Comparing the SSIM scores of our method with other three benchmarks on Subject 1, 2 and 3. [b] Comparing our method's 50-way Image and Video Classification Accuracy with previous SOTA model on Subject 1.}
\label{ssim-result}
\end{figure}

\section{Results}
\subsection{Video Reconstruction Performance}


\begin{figure}[h]
\centering
\small
\includegraphics[width=3.2in]{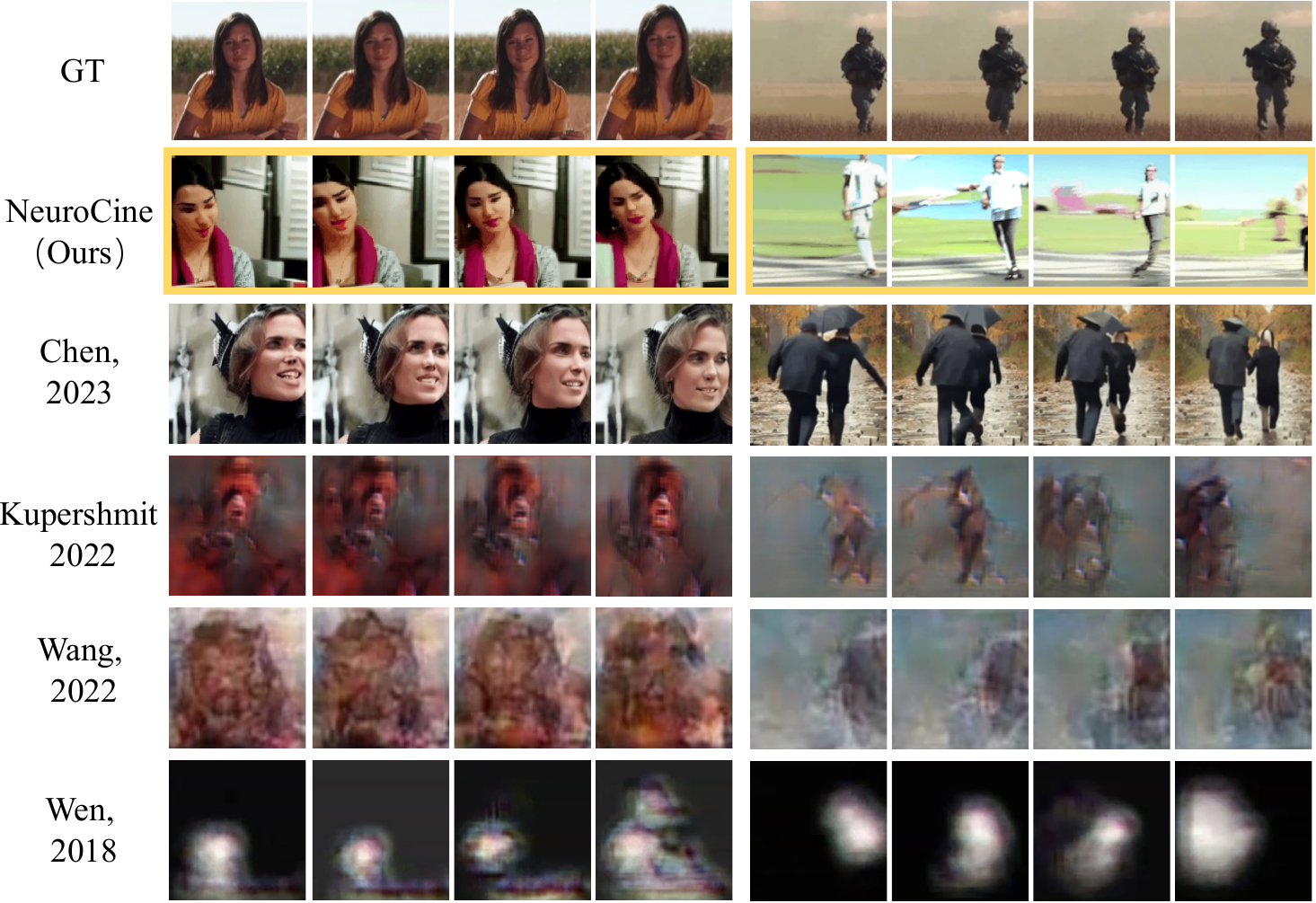}
\caption{Comparison of the decoded results considering our framework NeuralFlix and baselines.}
\label{vis-result3}
\end{figure}

\begin{figure}[h]
\centering
\small
\includegraphics[width=3.2in]{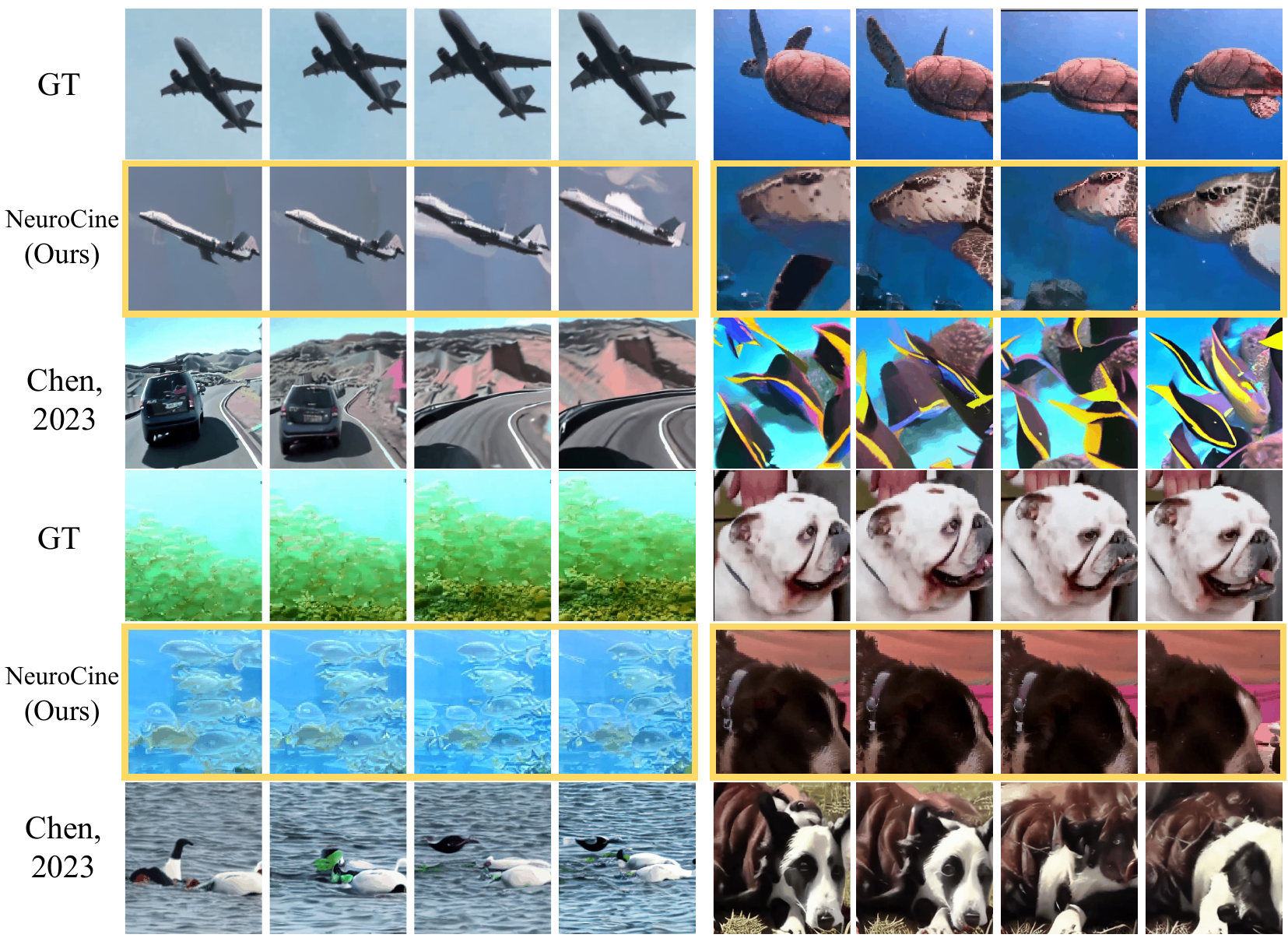}
\caption{ Additional comparisons of decoded results between our method NeuralFlix and the previous state-of-the-art (SOTA) model.}
\label{vis-result4}
\end{figure}

\begin{figure}[h]
\centering
\small
\includegraphics[width=3.2in]{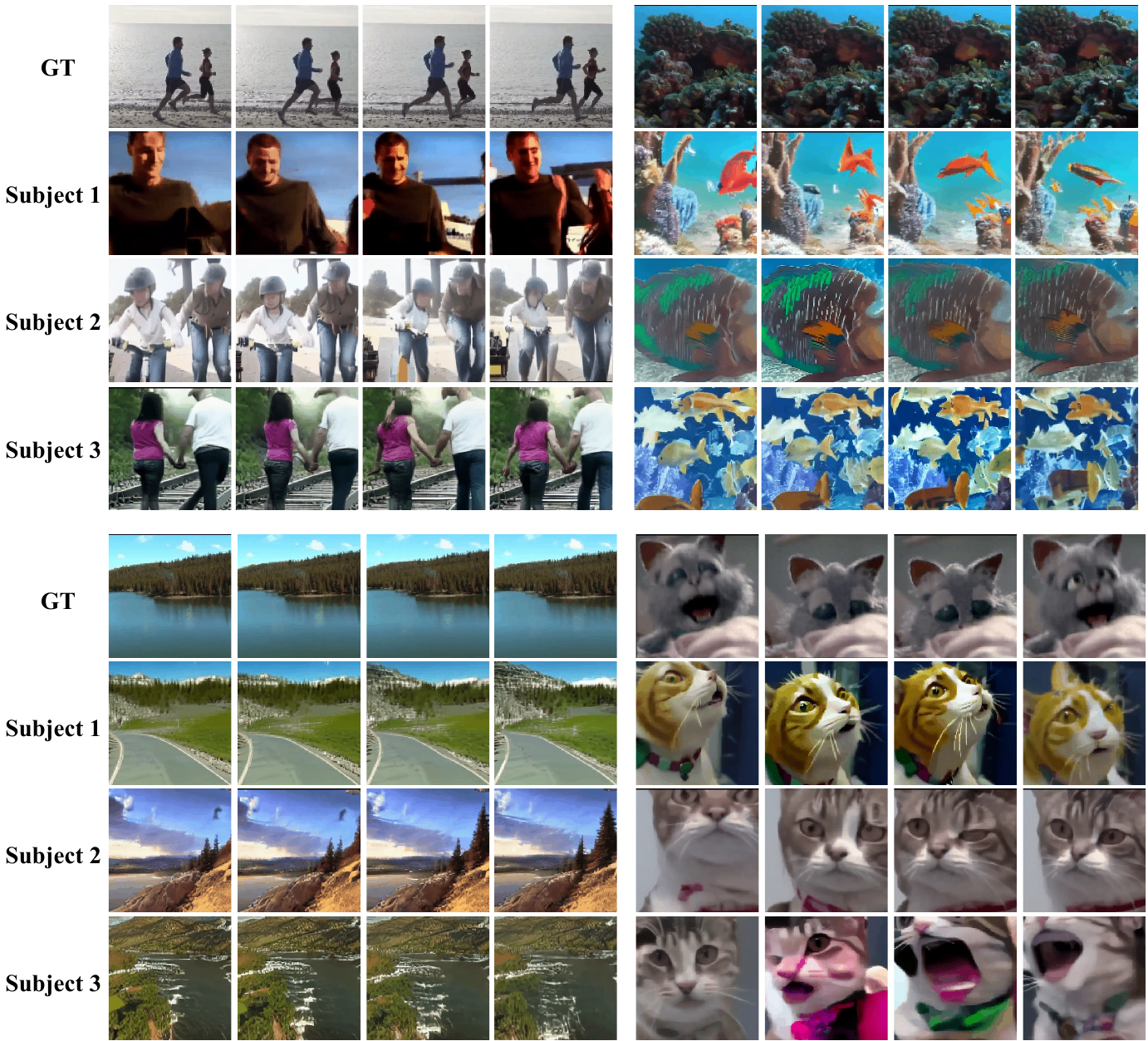}
\caption{Decoded results of our framework NeuralFlix from all three subjects in the dataset.}
\label{allsubs}
\end{figure}

\begin{table*}[h]
\centering
\small
\includegraphics[width=6.25in]{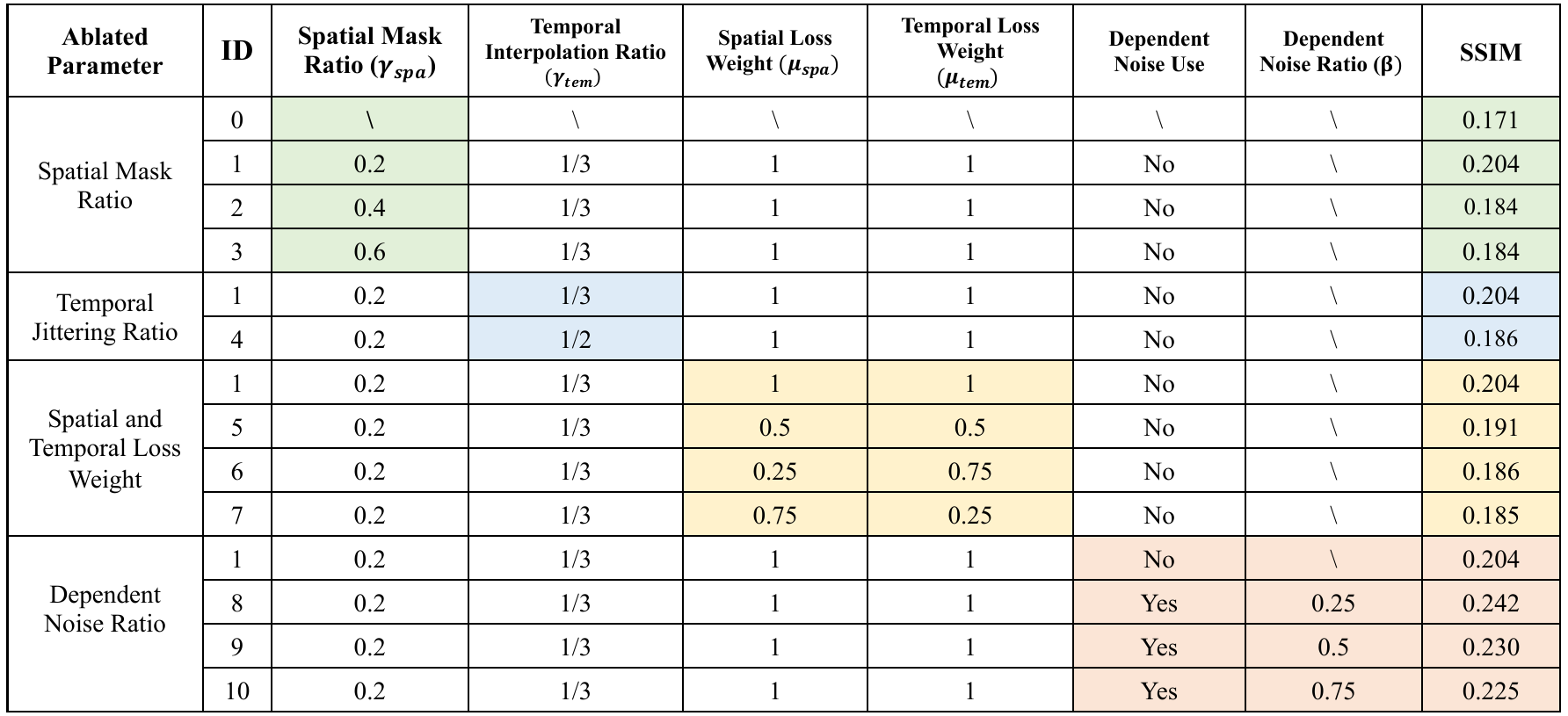}
\caption{Ablation study about NeuralFlix's important hyperparameters' effects on final video decoding performance measured by SSIM, including spatial mask ratio, temporal interpolation ratio, spatial loss weight, temporal loss weight, using of dependent noise and dependent noise ratio.}
\label{all_roi}
\end{table*}

\begin{figure*}[ht]
\centering
\small
\includegraphics[width=6.5 in]{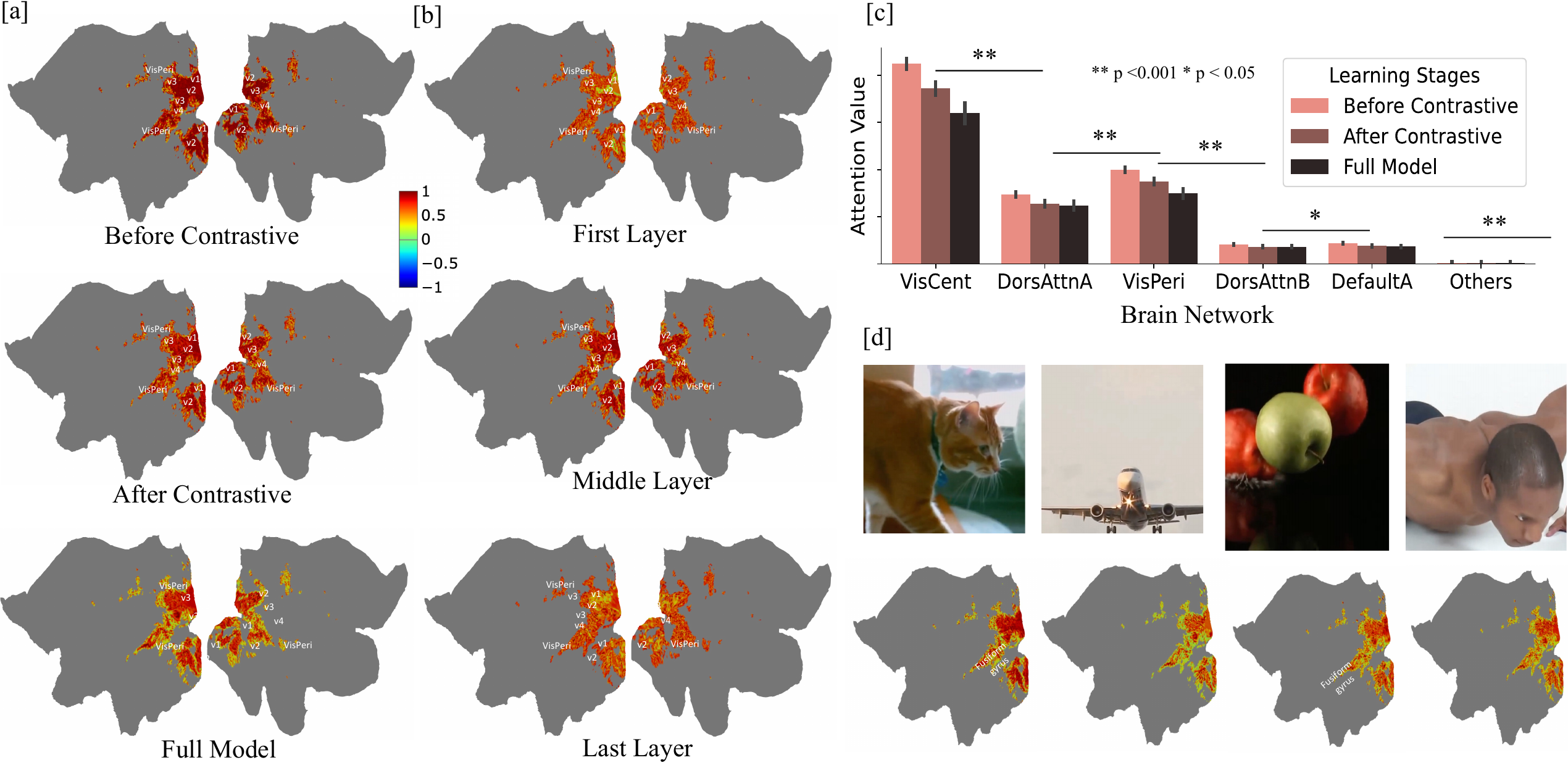}
\caption{Self-Attention Analysis for Subject 1. The gradation from blue to red on the voxel map signifies the ascending importance of each voxel in the generation process, as determined by the attention analysis. Panels [a] and [c] emphasize the evolution of attention maps throughout the learning stages. Panel [b] highlights the differential attention patterns across various transformer layers. Panel [d] presents a comparative view of attention maps corresponding to a few video frame examples under different categories.}
\label{interpretation}
\end{figure*}


In this section, we compare our methodology with baselines focusing on Structural Similarity Index Measure (SSIM) scores and classification accuracy. Our SSIM results, shown in Figure~\ref{ssim-result} [a], indicate our method sets a new state-of-the-art (SOTA) with scores of $0.225$, $0.224$, and $0.211$ for Subject 1, 2, and 3. Specifically, our method surpasses the previous (~\citet{chen2023cinematic}) by a significant margin of $\bold{20.97\%}$ in Subject 1, $\bold{31.00\%}$ in Subject 2 and $\bold{12.30\%}$ in Subject 3. (You may refer to hyperparamter settings in Table 1's experiment 10 for reproduction of Subject 1's decoding performance.) Figure~\ref{ssim-result} [b] further shows that our method achieves superior accuracy in both 50-way Image and Video Classification Accuracy. 

Qualitative comparisons in Figure~\ref{vis-result3} show that, unlike other models that yield blurry or unrecognizable outputs, our method and \citet{chen2023cinematic}'s approach produce high-quality, semantically accurate videos. A detailed comparison with the previous SOTA method (\citet{chen2023cinematic}) in Figure~\ref{vis-result4} reveals our superior semantic alignment with ground truth videos. For instance, where our method accurately generates a video of a swimming turtle, \citet{chen2023cinematic}'s output shows a group of fish. The versatility of our model is further demonstrated in Figure~\ref{allsubs} which shows our model's reconstructed results on all subjects, where it consistently decodes high-quality, semantically accurate videos from different subjects. We present more video frames generated by our model in Figure~\ref{appendix-vis} in the appendix. Considering that the static presentations of the video frames may not fully reflect how well our model can decode the dynamics of videos, we upload anonymously to FigShare\footnote[1]{https://figshare.com/s/dd158d4d530eb0816686} some gifs of the decoded video pieces for qualitative evaluation.


\subsection{Ablation Study}

In this subsection, we perform an ablation study on the validation set to assess the impact of each component within our model and the significance of hyperparameter settings on video decoding performance.

\textbf{Spatial and Temporal Augmentation:} The ratios for spatial masking and temporal interpolation dictate the extent of modification to the fMRI token vectors to generate augmented samples. Results from experiments [0-3] in Table 1 indicate that spatial augmentation significantly enhances decoding performance, with a spatial mask ratio of 0.2 yielding the best results. However, excessive masking detrimentally affects the SSIM of the reconstructed samples, underscoring the balance needed to utilize fMRI's spatial redundancy effectively. A similar balance is crucial for the temporal interpolation ratio, where a ratio of 1/3 enhances performance without the adverse effects seen when the ratio is increased further.

\textbf{Augmentation Loss Weight:} The augmentation loss weights determine their contribution to the total loss in optimizing the fMRI encoder. Experiments [1, 5-7] in Table 1 demonstrate that a balanced augmentation loss between spatial and temporal aspects improves performance, with setting both weights to 1 achieving the highest SSIM. This highlights the importance of evenly capturing spatial and temporal features of fMRI data for accurate video decoding.

\textbf{Dependent Noise Ratio:} Introducing dependent noise targets the inherent noise in fMRI data, enhancing the model's ability to decode videos, as shown in experiments [1, 8-10] in Table 1. This could confirm our hypothesis that incorporating dependent noise as a prior enhances video reconstruction with Diffusion Models. However, we also observe that increasing the noise ratio ($\sqrt{\beta}$) too much can negatively affect performance. This outcome is anticipated, since a higher noise ratio results in each frame becoming more similar to the others, leading to the creation of videos with frames that are too closely aligned, lacking variation and dynamism, and potentially resulting in nearly static videos. 

\textbf{Reliance on Diffusion Model Priors:} To evaluate whether our model learns to decode the dynamics of videos from the fMRI sequences but not purely rely on the prior of diffusion model to animate the images, we further use time-averaged brain signals to re-run the full pipeline (This part of results is not displayed Table 1).  This resulted in a significant decrease in performance. The 50-way-top-1 video classification accuracy on Subject 2 decreases from 0.187 ± 0.015 to 0.157 ± 0.013, and on Subject 3 from 0.195 ± 0.014 to 0.158 ± 0.012. The impact of time-averaging on fMRI is to prevent the model from predicting motion by utilizing temporal changes in brain signals. So the drops in metric confirm that our model's prediction of dynamics is attributable to its reliance on brain signals rather than a mere dependency on prior knowledge of diffusion models. It also demonstrates that our metrics effectively assess video dynamics.

\subsection{Interpretation of fMRI Encoder}

The results presented in Figure~\ref{interpretation} reinforce the vital role of the visual cortex in decoding visual spatiotemporal information, as previous literature \cite{zhou2013spatiotemporal,vetter2014decoding}.
The color gradation in the voxel map, transitioning from blue to red, represents a spectrum of attention weights, where red denotes areas with higher attention weights. These weights are derived from the self-attention mechanism of our fMRI encoder and emphasize the relative importance of each voxel in decoding the presented visual stimuli. 
For clarification, in Figure 6, brain networks are interconnected regions of the brain that collaborate to perform specific cognitive functions and processes.  We refer to the visual cortex networks (central and peripheral visual field; i.e. VisCent and VisPeri), dorsal attention networks (DorsAttnA and DorsAttnB), and the default mode network (DefaultA), within Yeo17 networks in the analysis, which are critical in processing spatio-temporal visual and semantic information, aligned with prior neuroscientific studies. 

As the fMRI encoder progresses through phases of contrastive learning and full training, as shown in Figure~\ref{interpretation} panels [a] and [c], there is a notable enhancement in the encoder's ability to process semantic information. This is evidenced by an increasing trend (p<0.001) to extract more semantic content from the visual stimuli, coupled with a shift in attention focus towards higher visual regions post-contrastive learning. Such evolution signifies the encoder's transition from primarily engaging with low-level visual features to interpreting high-level semantic features. This progression also results in a reduction in the total sum of attention values (p<0.05), indicating a more efficient and focused allocation of attention towards discriminative features, which enables a refined representation that encapsulates more relevant information with less cognitive load.

Furthermore, our exploration of attention maps in response to individual video stimuli substantiates the model's capability to distinguish between various visual categories and their associated cerebral patterns. This differentiation is exemplified in Figure~\ref{interpretation}[d], where increased attention values are observed in the fusiform gyrus when subjects view images of cats, in stark contrast to their neural responses to food-related images. Similarly, distinct attention patterns in the extrastriate body area (EBA) when viewing human images versus airplane images underline the EBA’s specialized function in processing biological forms.

\section{Conclusion}

In conclusion, this study introduces a novel dual-phase framework for decoding high-quality videos from fMRI data, effectively tackling challenges like spatial redundancy and temporal lags in fMRI signals. Our method, which combines spatial-temporal contrastive learning with an enhanced video diffusion model, shows notable improvements over existing models in both SSIM and semantic accuracy. The empirical results, benchmarked against prior works, demonstrate the efficacy of our approach. This research not only advances the field of neural decoding but also opens new avenues for exploration in neural imaging and cognitive neuroscience, with potential applications in understanding human cognition and developing assistive technologies for people with disabilities.

\section*{Ethical Statement}


The fMRI data employed in our training have undergone processing to ensure that they do not contain any information that could be directly traced back to individual participants. Furthermore, the original collection of this fMRI data adhered to strict ethical guidelines and reviews, as detailed in the respective source publications. 


\bibliography{mybibfile}

\newpage
\appendix
\onecolumn
\section{Visualization of Decoding Outcomes}
\begin{figure*}[h]
\centering
\small
\includegraphics[width=6.3in]{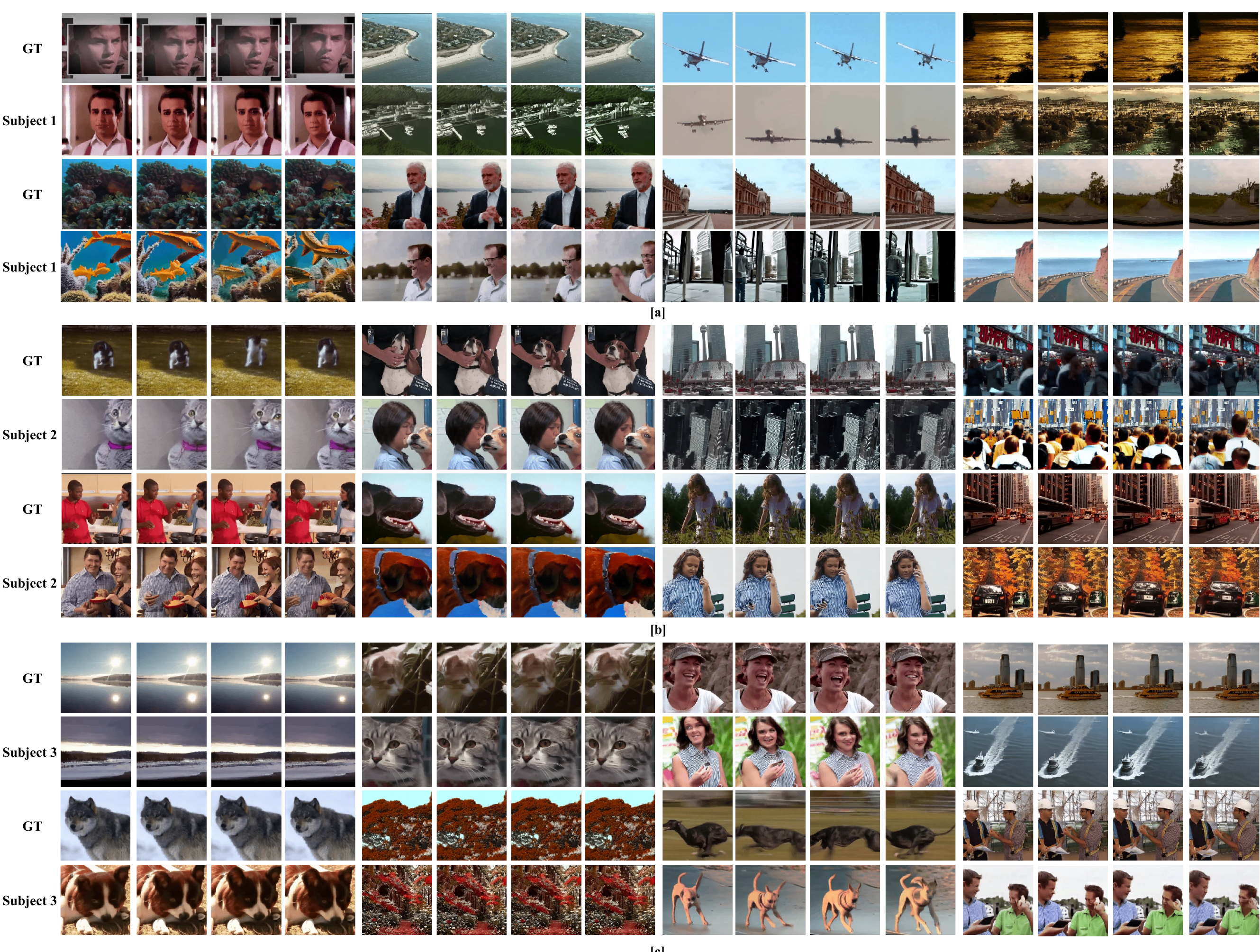}
\caption{Videos generated by our models: [a] Decoding outcomes for Subject 1, [b] Decoding outcomes for Subject 2, [c] Decoding outcomes for Subject 3.}
\label{appendix-vis}
\label{interpretation2}
\end{figure*}

In Figure~\ref{appendix-vis}, we showcase additional decoding videos for all three subjects examined in our study. These findings illustrate the capability of our proposed model to decode high-quality videos with precise semantics. Furthermore, the diverse collection of videos displayed below highlights the model's ability in decoding a broad range of semantic content from human brain activity. This includes videos related to objects, human motion, scenes, animals, et al.

\end{document}